\documentclass[a4paper]{./styles/svproc} %[a4paper]
\usepackage{url}
\usepackage{graphicx}
\usepackage{xcolor}
\usepackage{colortbl}
\usepackage{color}
\usepackage[colorlinks, linkcolor=pink]{hyperref}
\usepackage{multicol}
\usepackage{multirow}
\usepackage{amsfonts,amssymb}

\usepackage{array}
\usepackage{amsmath}
\usepackage{mathtools}

\begin{document}
\mainmatter              % start of a contribution
\title{DRAMA: An Efficient End-to-end Motion Planner for Autonomous \underline{Dr}iving with M\underline{am}b\underline{a}}
\titlerunning{DRAMA: Driving with Mamba}  % abbreviated title (for running head)
%                                     also used for the TOC unless
%                                     \toctitle is used
%
\author{Chengran Yuan\inst{1} \and Zhanqi Zhang\inst{1} \and Jiawei Sun\inst{1}
\and Shuo Sun\inst{1} \and Zefan Huang\inst{1} \and Christina Dao Wen Lee\inst{1} \and Dongen Li\inst{1} \and Yuhang Han\inst{1}
\and Anthony Wong\inst{2} \and \\ Keng Peng Tee\inst{2} \and Marcelo H. Ang Jr.\inst{1}
}
\authorrunning{Chengran Yuan et al.} % abbreviated author list (for running head)
%
%%%% list of authors for the TOC (use if author list has to be modified)
\tocauthor{Chengran Yuan, Zhanqi Zhang, Jiawei Sun, Shuo Sun,
Zefan Huang, Christina Dao Wen Lee, Dongen Li, Yuhang Han, 
Anthony Wong, Keng Peng Tee and Marcelo H. Ang Jr.}
\institute{National University of Singapore, Singapore\\
\email{\{chengran.yuan,zhanqizhang,sunjiawei,shuo.sun,huangzefan,\\christinaldw,li.dongen,yuhang\_han\}@u.nus.edu; \\mpeangh@nus.edu.sg},
\and
Moovita Pte Ltd, Singapore\\ %599489,
\email{\{anthonywong, kptee\}@moovita.com}
\\Project page: \href{https://chengran-yuan.github.io/DRAMA/}{https://chengran-yuan.github.io/DRAMA/}
}

\maketitle              % typeset the title of the contribution

\begin{abstract}
Motion planning is a challenging task to generate safe and feasible trajectories in highly dynamic and complex environments, forming a core capability for autonomous vehicles. In this paper, we propose DRAMA, the first Mamba-based end-to-end motion planner for autonomous vehicles. DRAMA fuses camera, LiDAR Bird's Eye View images in the feature space, as well as ego status information, to generate a series of future ego trajectories. Unlike traditional transformer-based methods with quadratic attention complexity for sequence length, DRAMA is able to achieve a less computationally intensive attention complexity, demonstrating potential to deal with increasingly complex scenarios. Leveraging our Mamba fusion module, DRAMA efficiently and effectively fuses the features of the camera and LiDAR modalities. In addition, we introduce a Mamba-Transformer decoder that enhances the overall planning performance. This module is universally adaptable to any Transformer-based model, especially for tasks with long sequence inputs. We further introduce a novel feature state dropout which improves the planner's robustness without increasing training and inference times. Extensive experimental results show that DRAMA achieves higher accuracy on the NAVSIM dataset compared to the baseline Transfuser, with fewer parameters and lower computational costs.

\keywords{Autonomous Driving, Motion planning, End-to-End, Navigation, Multi-modal}
\end{abstract}
\section{Introduction}
%

%%Backgroud

For decades, motion planning has been a heated research topic in the robotics field. As a core module in the autonomous driving stack, the motion planner is mainly responsible for generating a safe and feasible trajectory for the autonomous vehicle (AV) to execute in the future. However, it is challenging to obtain reliable and efficient trajectory planning due to factors including but not limited to accurate intention prediction of other road users, understanding of traffic signs and signals, encoding complexities of road topology, and intermediate reaction to other unforeseen obstacles and risks. 

Based on the different methods used, existing motion planners in autonomous vehicles can be categorized into two main groups: rule-based and learning-based. Conventional rule-based planners can perform consistently well in most driving scenarios but require deliberate fine-tuning for corner cases. In order to achieve improved scalability and generalizability, researchers have recently resorted to learning-based methods. Inspired by the prominent performance of Transformers in various motion prediction models \cite{liu_multimodal_2021,shi2022motion,zhou_query-centric_nodate,shi_mtr++_2023,sun2023get,sun2024controlmtr}, researchers explored employing Transformers with other foundation models such as CNN, RNN, LSTM, and  \cite{hu2023planning,huang2023gameformer,hu2023imitation,chitta2023pami} in motion planning tasks to leverage its superior capabilities in digesting topological map information and modeling the complex interactions among agents.

Although directly adopting methods from prediction tasks into planning tasks have been demonstrated plausible, these methods often suffer from intensive computation due to the quadratic cost of attention in Transformer-based backbones. To improve the inference speed and planning performance, researchers tried refining model architectures and proposing new training techniques, exploiting the potential of Transformer-based models. However, achieving more reliable, robust, and efficient planning capabilities demands a more powerful backbone, and new foundation models often provide models with higher potential. 
Recently, Mamba \cite{gu2023mamba,dao2024transformers} was proposed as a more advanced foundation model, which has demonstrated superior efficiency and accuracy in various downstream tasks \cite{zhu2024vision,wang2024mambayolossmsbasedyolo}. 

In this paper, we propose DRAMA, a Mamba-embedded encoder-decoder architecture for effectively fusing features from both camera and LiDAR bird's-eye-view (BEV) images and efficiently generating future trajectories. In the encoder architecture, we follow the fusion architecture proposed in \cite{chitta2023pami} and introduce multi-scale convolution, Mamba fusion to boost the model's effectiveness and efficiency. Inspired by concepts from the field of visual perception, multi-scale convolution blocks are designed to exploit scene information across multiple scales. We also show that Mamba \cite{dao2024transformers} introduced in our model is able to effectively self-attend to fused camera-LiDAR image features and achieve superior performance while reducing both model size and training cost. In the decoder architecture, the feature state dropout (FSD) and the Mamba-Transformer (MT) decoder are proposed to improve model robustness and performance. The FSD module can enhance the robustness and generalizability of the planner by mitigating the adverse effects of flawed sensor inputs and missing ego states, which may arise from data collection instability and inconsistencies. Meanwhile, the MT block is designed to perform self-attention within the vanilla Transformer decoder, which offers significant advantages compared to Transformer-based architectures.

\begin{figure*}[ht]
    \centering
    \includegraphics[width=\linewidth]{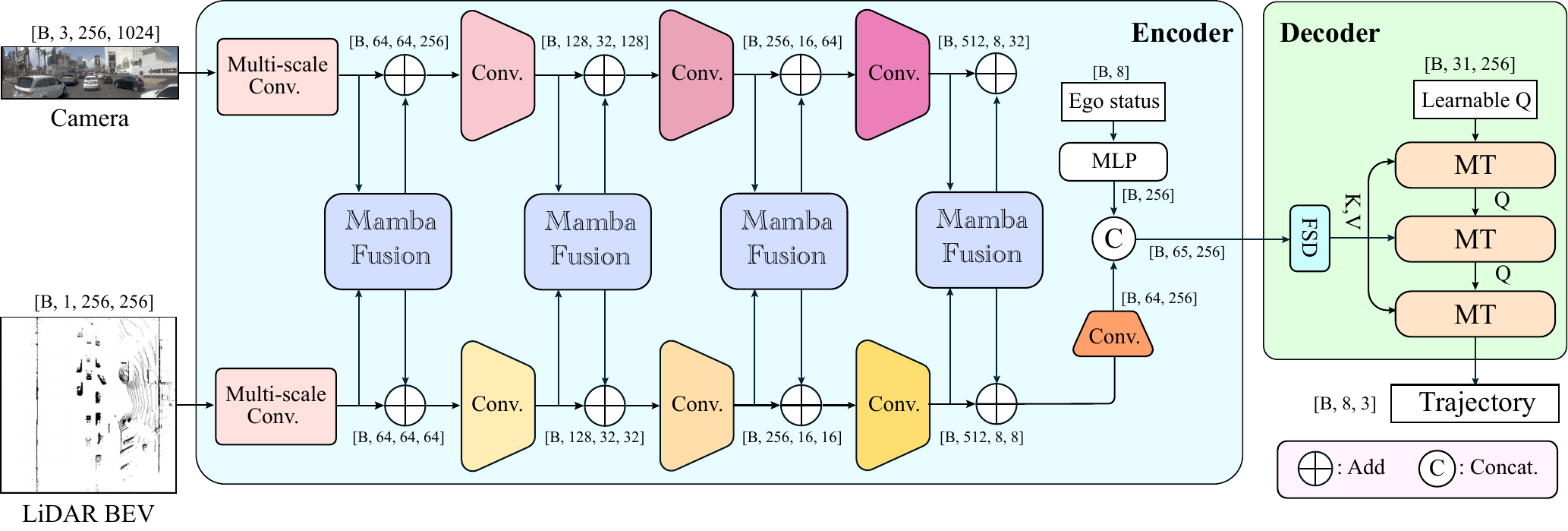} 
    \caption{\textbf{Pipeline} DRAMA combines camera and LiDAR BEV images in feature space using the Mamba Fusion module. The final fusion feature is concatenated with the ego status and passed into the decoder, which outputs a deterministic trajectory for AV navigation using multiple Mamba-Transformer decoder layers. }
    \label{fig:pipeline}
    \vspace{-1em}
\end{figure*}

As illustrated in Fig. \ref{fig:pipeline}, we first encode the cropped multi-view camera and LiDAR BEV images with the multi-scale convolution modules, after which encodings are reshaped, concatenated, and then fed into Mamba-2. The features processed by Mamba-2 are then split, reshaped, and added to the original features. We use four Mamba Fusion modules to fuse camera-LiDAR features. The final fused LiDAR BEV encoding is fed into the FSD module, and the features processed by FSD are later concatenated with ego status encodings as the encoder output. A learnable query and the encoder output are the Mamba-Transformer decoder layer inputs. After three MT layers, the model generates the future trajectory for the autonomous vehicle. The main contribution of this work can be summarised as follows:

1. We introduce a Mamba-embedded encoder-decoder architecture named DRAMA, which includes an encoder that effectively fuses features from the camera and LiDAR BEV images with a Mamba Fusion module and the decoder generates a deterministic trajectory with the Mamba-Transformer decoder, which is universally adaptable for any Transformer-based model.

2. We incorporate multi-scale convolution and feature state dropout modules with a differentiated dropout policy in DRAMA. These modules improve the model's effectiveness and robustness by extracting scene information at multiple scales and mitigating the effects of noisy sensor input and missing ego states. 
    
3. The proposed modules and overall architecture were evaluated using the NAVSIM planning benchmark \cite{dauner2024navsim}. Experimental results demonstrate that our model achieves notable performance improvement while using fewer model parameters and less training costs compared to the baseline.

\section{Related Works}

\subsection{Motion Planning for Autonomous Driving}
%% Conventional Planning

Motion planning for autonomous driving has been a long-researched topic in the robotic field. From the conventional perspective, motion planning is a downstream task of behavior planning or decision-making, and it is responsible for generating a drivable and comfortable trajectory guaranteeing safety \cite{paden2016survey}. Conventional motion planning usually relies on geometry and optimization and can be broadly classified into graph-based, sampling-based, and optimization-based approaches. Graph-based methods like A* \cite{Astar} and Hybrid A* \cite{Montemerlo2009} search for a minimum-cost path after discretizing the vehicle configuration space. Sampling-based methods like \cite{Howard2008,Werling2010,McNaughton2011,Xu2012,Gu2013,sun2022fiss,sun2023fiss+} create trajectory samples within the state or action space to discover feasible paths. In contrast, optimization-based methods employ techniques such as the EM algorithm \cite{Fan2018}, and convex optimization \cite{Ziegler2014} to determine an optimal trajectory that meets specified constraints. These methods typically involve extensive manual design and optimization and often lack generalizability in dynamic or changing environments.

%% Learning Based Planning 
%% nuPlan & Navsim
Learning-based trajectory planning has been significantly accelerated with the release of public driving datasets and benchmarks specifically for motion planning purposes. Currently, nuPlan \cite{caesar2021nuplan} is the largest annotated planning dataset and benchmark for motion planning. Based on the nuPlan and OpenScene \cite{contributors2023OpenScene} dataset, a recent dataset called NAVSIM \cite{dauner2024navsim} is developed to address the misalignment between open-loop and close-loop evaluation metrics and serve as a middle ground between these evaluation paradigms. 

Based on these open-source datasets, \cite{dauner2023parting} analyzed misconceptions in data-driven motion planning methods and proposed a simple yet efficient planner, which ranked first on the nuPlan leaderboard. However, this planner is highly optimized for the nuPlan metrics \cite{huang2023dtpp}, and its performance degrades when transferred to other scenarios. \cite{hu2023imitation,huang2023gameformer} both employed a post-optimizer to refine the trajectory, but the imitations of these solvers arise from unstable inference speed and reliance on initial guesses. \cite{cheng2024pluto} introduced an architecture to simultaneously plan the lateral and longitudinal trajectory, whose post-processing module may fail if trajectories sampled by the planner are unstable. \cite{hu2023planning} devised a planning-oriented end-to-end autonomous driving framework, while the architecture is not lightweight and results in extensive computation cost. The best solution of NAVSIM \cite{li2024hydramdp} introduced a teacher-student knowledge distillation pipeline where the student model learns a range of trajectory candidates tailored to various evaluation metrics from both human and rule-based teachers, but this distillation intensifies the computation. These existing learning-based methods tend to over-emphasize the metric performance, often at the expense of computational efficiency. Many of these approaches suffer from unbearable computational burdens due to the complex architectural design or online simulation for trajectory scoring and refinement. To alleviate computational intensity and enhance performance, we propose DRAMA, a Mamba-embedded encoder-decoder pipeline designed to achieve efficient and superior planning performance.

% \textcolor{red}{To deal with the misalignment between open-loop and close-loop evaluation, based on OpenScene \cite{contributors2023OpenScene} and nuPlan dataset, the NAVSIM dataset \cite{dauner2024navsim} is developed to serve as a middle ground between these evaluation paradigms.} 

% multi-scale convolution

% Masked-autoencoders

\subsection{State Space Models}

To mitigate the substantial computational and memory demands of State Space Models (SSMs) when modeling long-range dependencies, \cite{gu2021efficiently} proposed the structured state-space sequence models (S4), which modifies the $\mathbf{A}$ matrix in SSM into a conditioning matrix with a low-rank correction. \cite{gu2023mamba} introduced an input selection mechanism and a hardware-aware parallel algorithm, which improve the performance of S4 and significantly increase computational efficiency. This enhanced model, Mamba, demonstrates substantial application potential in image processing \cite{zhu2024vision}, language processing \cite{he2024densemambastatespacemodels,lieber2024jambahybridtransformermambalanguage,pioro2024moe} and other domains \cite{zhang2024motionmambaefficientlong}. \cite{dao2024transformers} theoretically demonstrate the equivalence of SSMs to semiseparable matrices. Additionally, the State Space Duality (SSD) is introduced to enhance the original Mamba, and the design incorporates multi-head attention (MHA) to SSMs to optimize the framework, resulting in the improved version (Mamba-2) exhibiting greater stability and improved performance. Inspired by the previous success of the Mamba family, we apply the up-to-date architecture, Mamba-2, to end-to-end motion planning. To the best of our knowledge, this represents the first application of the Mamba-2 in the field of autonomous driving. For the sake of clarity and brevity, unless otherwise specified, all subsequent references to Mamba pertain to Mamba-2.

% \section{Problem Formulation}

\section{Our Method}
We present our Mamba-based end-to-end motion planning framework, DRAMA, which utilizes convolutional neural networks (CNNs) and Mamba to encode and fuse features from the camera and LiDAR BEV images. The decoder employs our proposed Mamba-Transformer decoder layer to decode the final trajectory. In the following sections, we will explore our four designed modules in detail: the Mamba fusion block, the Mamba-Transformer decoder layer, multi-scale convolution, and feature state dropout.

\subsection{Mamba Fusion Block and Mamba-Transformer}

\subsubsection{Mamba Preliminaries}

Structured state space sequence models (S4), derived from the continuous system,  utilize a 1-D input sequence or function $x(t) \in \mathbb{R}$ and intermediate hidden state  $h(t) \in \mathbb{R}^N $ to produce the final output $y(t) \in \mathbb{R}$. The intermediate hidden state $h(t)$ and input $x(t)$ are employed to compute the $y(t)$ via projection matrices $\mathbf{A} \in \mathbb{R}^{N \times N}$, $\mathbf{B} \in \mathbb{R}^{N \times 1}$, and $\mathbf{C} \in \mathbb{R}^{1 \times N}$

\begin{equation}
    \begin{array}{rcl}
        h^{\prime}(t) &=& \mathbf{A} h(t)+\mathbf{B} x(t), \\
        y(t) &=& \mathbf{C} h(t).
    \end{array}
\label{eq:1}
\end{equation}

% \begin{figure*}[ht]
%     \centering
%     \includegraphics[width=3cm]{Drama/figs/Mamba.pdf} 
%     \caption{Mamba Block.}
%     \label{fig:Mamba}
%     \vspace{-1em}
% \end{figure*}

The system applies the learnable step size $\Delta$ and zero-order hold to transform the continuous system into a discretized one. Consequently, Eq. (\ref{eq:1}) can be reformulated as follows: 

\begin{equation}
    \begin{array}{rcl}
         h_t &=& \overline{\mathbf{A}} h_{t-1}+\overline{\mathbf{B}} x_t, \\
         y_t &=& \mathbf{C} h_t.
    \end{array}   
    \label{eq:discretized}
\end{equation}

By mathematical induction, the final output of Eq. (\ref{eq:discretized}) can be rewritten as follows:
\begin{equation}
    \begin{array}{rcl}
        y_t &=& \sum_{s=0}^t C_t^\top A_{t:s}^\times B_s x_s, \\
        y &=& \text{SSM}(A, B, C)(x) = Mx, \\
        % M_{ji} &\coloneqq& C_j^\top A_j \cdots A_{i+1} B_i
    \end{array}   
\end{equation}
and the matrix $M$ is defined as follows:
\begin{equation}
    \begin{array}{rcl}
       M_{ji} &\coloneqq& C_j^\top A_j \cdots A_{i+1} B_i
    \end{array}   
\label{eq:ssm}
\end{equation}
where $A_{t:s}^\times$ denotes the product of the matrices from $A_{t}$ to $A_{s}$, and the indices $j$ and $i$ denote the $j$-th and $i$-th $A$, $B$, $C$ matrices, respectively.

The lower triangular SSM transformation matrix $M$, as described by Eq. (\ref{eq:ssm}) also satisfies the definition of $N$-sequentially semiseparable $(SSS)$ representation. Consequently, the SSM and SSS representations are equivalent. 

As a result, structured matrix multiplication for $SSS$ can be effectively utilized in computations involving $SSM$. To implement this method, parameter matrix $M$ is decomposed into a diagonal block and low-rank block using the \textbf{Structured Masked Attention (SMA) quadratic mode algorithm} and the \textbf{SMA linear mode algorithm} \cite{dao2024transformers}, respectively. Furthermore, Multi-Head Attention (MHA) is employed to enhance model performance. 

\subsubsection{Mamba Fusion}

In order to capture multi-scale context on different modalities, the previous baseline \cite{chitta2023pami} implemented the self-attention layer in the Transformer to fuse and exploit features from the LiDAR and cameras. First, features from two modalities are transformed and concatenated, generating the combined feature $I$. Then the $I$ multiply three different projection matrix $M^Q$, $M^K$,  and, $M^V$ to get $Q$, $K$ and $V$. The final output of the fusion module can be computed by:

\begin{equation}
    Fusion = \text{softmax}\left(\frac{Q K^T}{\sqrt{D_k}}\right) V
\end{equation}

The overall training computational complexity is given by :

\begin{equation}
    \Omega(\text{self-attention}) = T^2D
\end{equation}
where $T$ and $D$ represent the length and dimension of the input, respectively.

\begin{figure*}[ht]
    \centering
    \includegraphics[width=10cm]{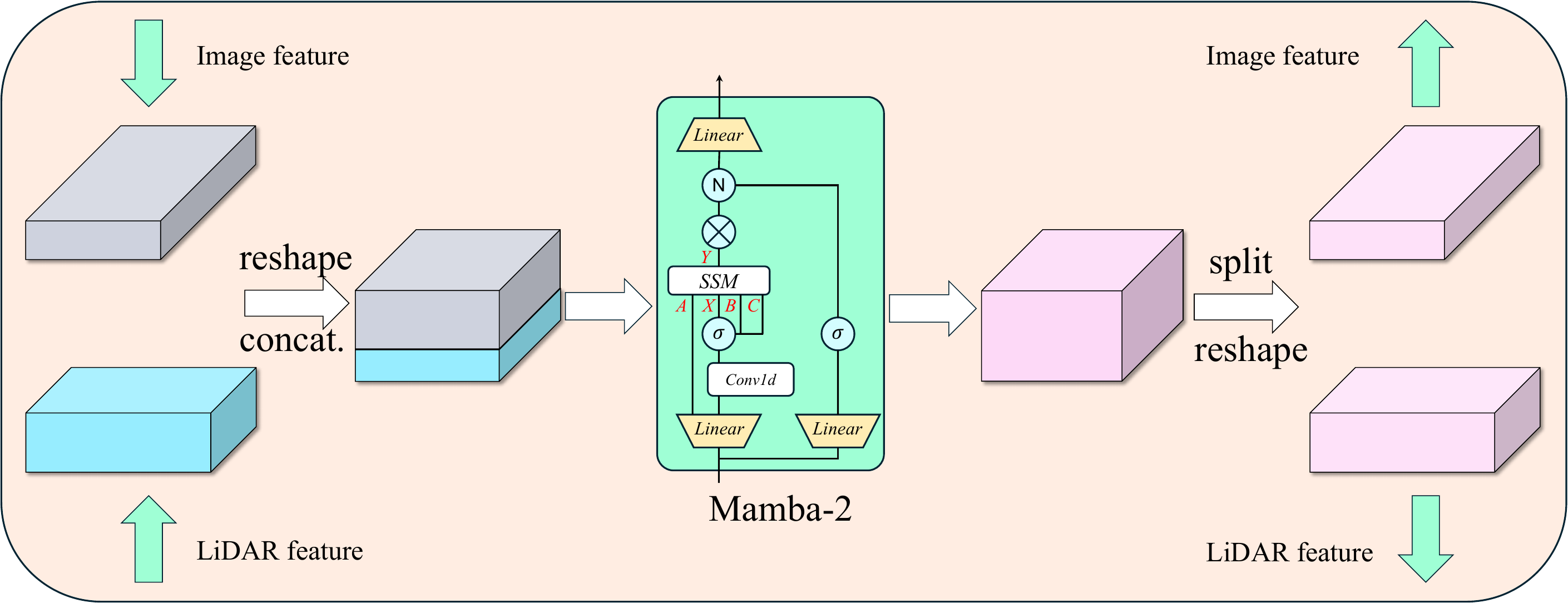} 
    \caption{Mamba Fusion. The image and LiDAR BEV features are initially reshaped and concatenated, followed by fusion through the Mamba module. After the processing by Mamba, the fused features are split and reshaped to their original dimensions.}
    \label{fig:MambaFusion}
    \vspace{-1em}
\end{figure*}

We propose utilizing Mamba as an alternative to self-attention for feature fusion due to its efficient matrix computations. We adhere to the fusion method implemented in \cite{chitta2023pami}, as illustrated in Fig. \ref{fig:MambaFusion}. Unlike \cite{chitta2023pami}, we utilize Mamba-2 instead of Transformer to process the fused features. The computational cost of Mamba is significantly reduced owing to the absence of complex computations present in traditional transformer self-attention. Assuming the head dimension $P$ is equal to the state dimension $D$, i.e., $P = D$, the training cost is given by:

\begin{equation}
    \Omega(\text{SSD}) = TD^2
\end{equation}

In our Mamba Fusion module, we set \(T = 320\) and \(P = 16\), theoretically achieving an approximately 20-fold reduction in training cost during the fusion process compared to self-attention.

\subsubsection{Mamba-Transformer Decoder}

\begin{figure*}[ht]
    \centering
    \includegraphics[height=2.5cm]{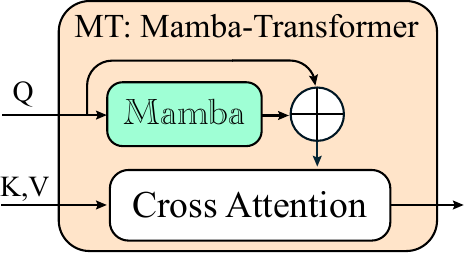} 
    \caption{Mamba Transformer Decoder. The query undergoes processing within the Mamba module and then engages cross-attention with the key and value.}
    \label{fig:Mamba-Transformer}
    \vspace{-2em}
\end{figure*}

As shown in Fig. (\ref{fig:Mamba-Transformer}), we combine Mamba and Transformer architectures to develop the novel Mamba-Transformer (MT) decoder. Initially, the learnable Query is passed into the Mamba component of the MT, which functions similarly to self-attention. As cross-attention with Mamba remains underexplored, we employ the Transformer cross-attention mechanism to attend to the query from Mamba along with the key and value from the FSD module.

\subsection{Multi-scale Convolution}

\vspace{-1em}
\begin{figure*}[ht]
    \centering
    \includegraphics[height=2.5cm]{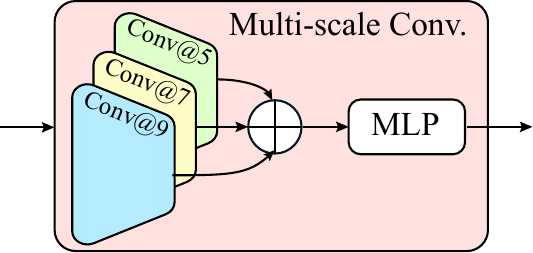} 
    \caption{Multi-scale Convolution.}
    \label{fig:MSC}
    \vspace{-2em}
\end{figure*}

To capture multi-scale image features, we utilize a multi-convolution design, as shown in Fig. (\ref{fig:MSC}), where the image is processed through three convolutional layers with varying kernel sizes—5, 7, and 9, respectively. The outputs from these convolution layers are combined and further encoded by a multi-layer perceptron (MLP) layer to enhance the model's perceptual capabilities.

\subsection{Feature State Dropout}
\begin{figure*}[ht]
    \centering
    \includegraphics[height=3cm]{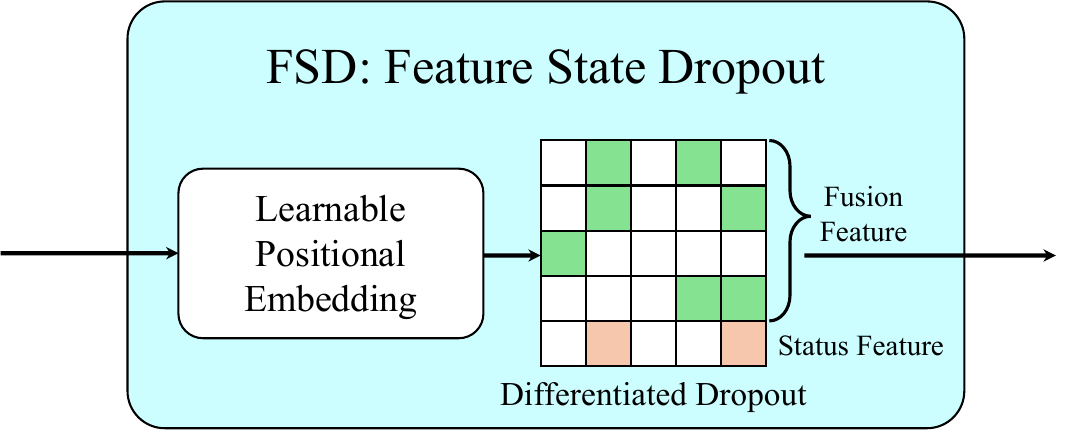} 
    \caption{Feature State Dropout. The fusion feature and state feature are concatenated and augmented with a learnable positional embedding. The combined features are then processed using the differentiated dropout policy to selectively drop features.}
    \label{fig:FSD}
    \vspace{-1em}
\end{figure*}

Due to hardware limitations and noise in onboard sensors, observations and perceptions of the surrounding environment (e.g., positions or velocities) may be inaccurate and may not fully reflect true conditions. Additionally, when driving commands from the navigation module are missing, or when navigating complex traffic conditions where human directives may be suboptimal, it becomes crucial for the model to deeply understand and reason about the scenarios and surrounding agents, even in the absence of explicit guidance. Previous studies \cite{he:cvpr,cheng2023plantf} have demonstrated that masking certain image and vehicle state features can enhance overall performance in self-supervised tasks and motion planning. To address these issues and build on these insights, we implement feature state dropout for image feature fusion from two modalities and the ego status, as illustrated in Fig. \ref{fig:FSD}. Initially, the features to be encoded are added with a learnable positional embedding, followed by the differentiated dropout to mask some features.

% \textcolor{red}{ To address this, we apply FSD to the fusion information to ensure that the model can analyze and understand the environment effectively even when sensory input is vague or partially accurate.}

% \textcolor{red}{
% Similarly, we also apply dropout to the ego status. For instance, when driving commands from the navigation module are missing, this situation simulates scenarios in complex traffic conditions where clear directives may be insufficient. By relying solely on the size and orientation of speed, acceleration, and the surrounding environment, this approach compels the model to deeply understand and reason about agents and scenarios, even in the absence of explicit guidance.}

We employ a differentiated dropout policy in DRAMA, which applies distinct dropout rates to the fusion and ego status features. A relatively low dropout rate is assigned for the fusion feature to preserve its integrity. This measure aims to avoid the excessive loss of fused perception information degrading overall performance.

\section{Experiment}

\subsection{Experimental Setup}
\subsubsection{Dataset}
We evaluate our models on NAVSIM dataset \cite{dauner2024navsim} developed on OpenScene\cite{contributors2023OpenScene}, which is a compact redistribution of the nuPlan dataset \cite{caesar2021nuplan}. OpenScene provides 120 hours of driving log, recorded at a lower frequency of 2Hz. The input data for the agent includes eight $1920 \times 1080$ pixel images from multiple views and a merged LiDAR point cloud generated from five sensors. The data input includes the current and, optionally, the previous three time-step frames, covering a span of 1.5 seconds at 2Hz.

\subsubsection{Metrics}
NAVSIM provides the Predictive Driver Model Score (PDMS) proposed in \cite{dauner2023parting} to evaluate the model performance, which is formulated as follows:

\begin{equation}
\text{PDMS}= \text{NC} \times \text{DAC} \times \frac{(5 \times \text{EP} + 5 \times \text{TTC} + 2 \times \text{C})}{12},
\end{equation}
where driving with no collision (NC) with traffic participants and no infraction regarding Drivable Area Compliance (DAC) are two hard penalties, and failing to meet this requirement will result in a PDMS of 0. The three sub-scores, namely Ego Progress (EP), Time-to-collision (TTC), and Comfort (C), are introduced to evaluate specific aspects of the planned trajectory. Ego progress assesses the ratio of ego progress along the route center, TTC measures the safety margins relative to other vehicles, and comfort evaluates the ride quality by comparing the acceleration and jerk of the trajectory to predetermined thresholds. We follow the implementation of these metrics as provided in NAVSIM benchmark \cite{dauner2024navsim}.

\subsection{Implementation Details}
We train, validate, and test our models with the standard dataset (all\_scenes) using an NVIDIA RTX3090Ti GPU, with a total batch size of 32 across 30 epochs.
For the optimizer, we use AdamW with learning rate and weight decay set to 1e\--4
and 0.01. Camera images from the front-left and front-right views are center-cropped and then concatenated with the front-view image to create a 256 × 1024 pixel image input. The LiDAR BEV image is generated by splatting the LiDAR points onto the BEV plane. We utilize only the current time-step camera and LiDAR images as input without incorporating any previous frames or applying data augmentations. Our input data also incorporates the current ego vehicle's status, including velocity, acceleration, and driving commands from the navigation module, such as turning, lane changing and following. The final output consists of an 8-waypoint trajectory over 4 seconds, sampled at 2 Hz, with each waypoint specified by x, y, and heading coordinates.

% \subsection{Main Results}
\subsection{Quantitative results}
The evaluation of the proposed modules against the baseline Transfuser (T) shows significant improvements in various metrics, as illustrated in Table 1. Integrating Multi-Scale Convolution (MSC) results in an enhanced PDM score, increasing from 0.835 to 0.843, highlighting its effectiveness in capturing multi-scale features that improve overall model performance. The inclusion of Mamba Fusion (MF) further boosts the PDM score to 0.848, with notable improvements in Ego Progress (EP) from 0.782 to 0.798, indicating superior modality fusion. Feature state dropout (FSD) shows the highest individual module enhancement for EP, achieving 0.802 and a PDM score of 0.848, demonstrating its role in mitigating poor sensor inputs. Additionally, the Mamba-Transformer (MT) module achieves a PDM score of 0.844, with a significant improvement in Time-to-Collision (TTC), underscoring its robust self-attention mechanism. The combination of these modules in DRAMA without MSC, namely T+MF+FSD+MT, leads to a PDM score of 0.853, with consistent improvements across all metrics, and the full DRAMA model achieves the highest PDM score of 0.855, confirming the effectiveness of the integrated approach.
\begin{table}[htbp]
\caption{Effectiveness Evaluation of the Proposed Modules}

\label{tab: Effectiveness Evaluation}
\centering
\resizebox{\textwidth}{!}{ % 调整表格大小
\begin{tabular}{c|ccccccc}
    \hline
    \rowcolor[rgb]{0.863,0.863,0.863} & ~~~~~~~Method~~~~~~~ & ~~ NC $\uparrow$ ~~& ~~DAC $\uparrow$~~ & ~~ EP $\uparrow$ ~~& ~~ TTC $\uparrow$~~& ~~C $\uparrow$~~ & PDM Score $\uparrow$ \\ 
    \hline
    \\[-1em]
    \small
    ~Baseline~ & Transfuser(T) & 0.975 & 0.916 & 0.782 & 0.935 & 1 & 0.835 \\
    \hline
    \\[-1em]
    \multirow{6}{*}{~Ours~} & T + MSC & 0.975 & 0.923 & 0.787 & 0.941 & 1 & 0.843 \\
        & T + MF & 0.975 & 0.923 & 0.798 & 0.944 & 1 & 0.848 \\
        & T + FSD & 0.978 & 0.926 & \textbf{0.802} & 0.942 & 1 & 0.848 \\
        & T + MT & 0.977 & 0.924 & 0.785 & 0.947 & 1 & 0.844 \\
        & DRAMA w/o MSC & 0.977 & 0.928 & 0.800 & \textbf{0.949} & 1 & 0.853 \\
        & DRAMA w/ MSC&\textbf{0.980} & \textbf{0.931} & 0.801 & 0.948 & 1 & \textbf{0.855} \\
    \hline
\end{tabular}
}  % 调整表格大小的结束括号
\end{table}

Table \ref{tab:FSD ablation} presents the impact of different feature state dropout rates on model performance, indicating that varying the dropout rates for state and fusion features can enhance the model's robustness and accuracy. The baseline Transfuser (T) achieves a score of 0.835. Introducing FSD with a fusion dropout rate of 0.1 improves the score to 0.842, and a state dropout rate of 0.5 yields a higher score of 0.844, suggesting that the model benefits from handling missing state features. The combination of a state dropout rate of 0.5 and a fusion dropout rate of 0.1 achieves the highest score of 0.848, indicating that balanced dropout rates between these two feature types optimize model performance. 

%%%%% 长度？
\begin{table}[htbp]
\caption{FSD Dropout Rate and Performance Comparison}
\label{tab:FSD ablation}
\centering
% \resizebox{\textwidth}{!}{ % 调整表格大小
\begin{tabular}{ccccccc}
    \hline
    \rowcolor[rgb]{0.863,0.863,0.863} ~~~~~~~Method~~~~~~ & ~~State Dropout Rate~~& ~~Fusion Dropout Rate~~ & ~~ PDM Score $\uparrow$ ~~ \\ 

    \hline
    \\[-1em]
    Transfuser (T) & 0 & 0 & 0.835 \\
    \hline
    T+ FSD & 0 & 0.1 & 0.842 \\
    T+ FSD & 0.5 & 0 & 0.844 \\
    T+ FSD & 0.5 & 0.1 & \textbf{0.848} \\
    \hline
\end{tabular}
% }  % 调整表格大小的结束括号
\end{table}

Table 3 provides a comprehensive comparison of the training and validation performance across various methods, emphasizing the efficiency of the proposed modules. The baseline Transfuser (T) has a total parameter size of 56 MB, with training and validation speeds of 4.61 iterations per second (it/s) and 9.73 it/s, respectively. Introducing the Multi-Scale Convolution (MSC) module slightly reduces the training speed to 3.77 it/s while maintaining a similar validation speed, indicating a trade-off between enhanced feature extraction and computational cost. Conversely, the Mamba Fusion (MF) module significantly reduces the total parameter size to 49.9 MB and improves the training speed to 4.92 it/s and validation speed to 9.94 it/s, showcasing its superior efficiency in modality fusion.

\begin{table}[htbp]
\caption{Efficiency Evaluation of Our Proposed Methods}
\label{tab: Efficiency Evaluation of Our Proposed Methods}
\centering
\resizebox{\textwidth}{!}{ % 调整表格大小
\small  % 调整表格字体大小
\begin{tabular}{c|ccccc}
    \hline
    \rowcolor[rgb]{0.863,0.863,0.863} ~ &  & ~Total Params~ & ~Training Speed~ & ~Validation Speed~ &~PDM~~~\\[-0.05em] 
    
    \rowcolor[rgb]{0.863,0.863,0.863} &\multirow{-2}{*}{Method}& (MB) $\downarrow$  & (it/s) $\uparrow$ & (it/s) $\uparrow$ & Score $\uparrow$  \\
    \hline
    \\[-1em]
    ~Baseline~&\textbf{Transfuser(T)} & 56  & 4.61 & 9.73 & 0.835\\
    \hline
    \\[-1em]
    \multirow{6}{*}{~Ours~} & T + MSC & 56.2  & 3.77 & 9.69 & 0.843\\
    &T + MF &  \textbf{49.9}  & \textbf{4.92} & \textbf{9.94} & 0.848 \\
    &T + FSD & 56  & 4.61 & 9.74 & 0.848 \\
    &T + MT & 56.5  & 4.51 & 9.75 & 0.844 \\
    &DRAMA w/o MSC & 50.4  & 4.84 & 9.91 & 0.853\\
    &DRAMA w/ MSC & 50.6 &  3.86 & 9.73 & \textbf{0.855} \\
    \hline
\end{tabular}
}  % 调整表格大小结束
\end{table}

The integration of feature state dropout (FSD) maintains speeds comparable to the baseline, demonstrating its efficiency without adding computational overhead. This finding highlights the universal and lightweight nature of the FSD module, which can be effectively incorporated into various models to enhance their performance.

The Mamba-Transformer (MT) module achieves a balanced improvement in both performance and speed, although it slightly reduces training speed to 4.51 it/s. This reduction is attributed to our input length $T=31$ being smaller than the state dimension $D=128$, which consequently increases the training cost from $\Omega(T^2D)$ to $\Omega(TD^2)$. The combined DRAMA architecture without MSC, further enhances efficiency by reducing the total parameters to 50.4 MB, with training and validation speeds of 4.84 it/s and 9.91 it/s, respectively. Finally, the full DRAMA model, incorporating all modules, maintains a reduced parameter size of 50.6 MB but experiences a slight decrease in training speed to 3.86 it/s. Despite this, it achieves the highest PDM score, validating the overall effectiveness and efficiency of the integrated approach.

\begin{figure*}[ht]
    \centering
    \includegraphics[width=12cm]{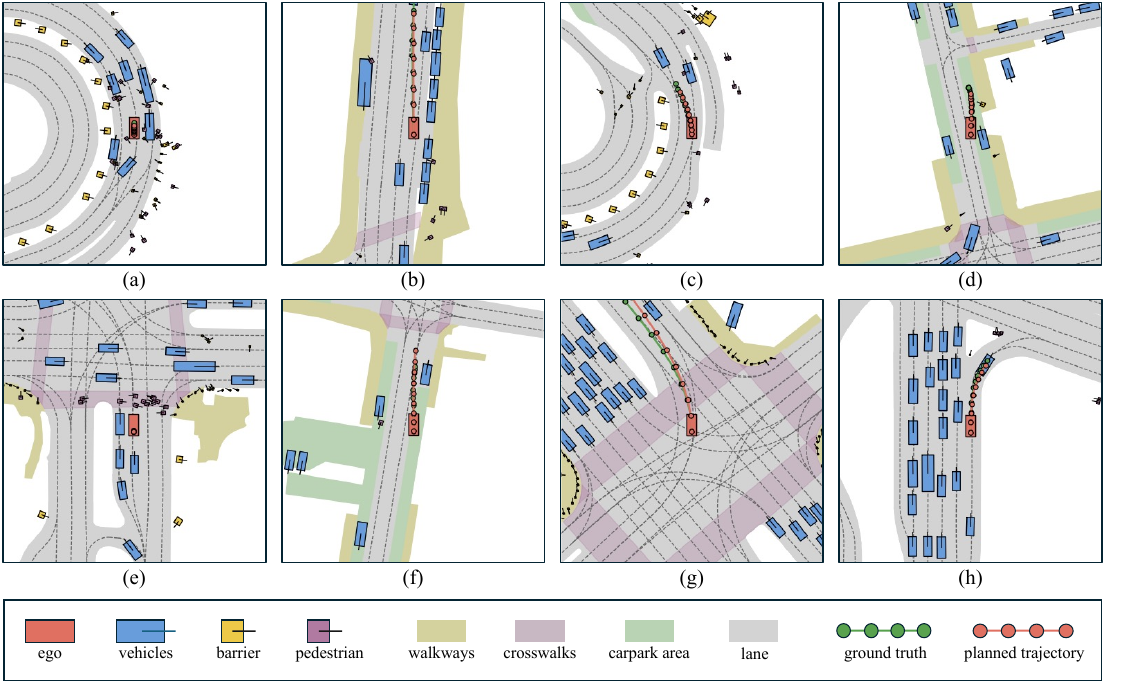}

    \caption{Visualizations of the planning results of DRAMA across various scenarios. (a) Yielding to pedestrians (b) Lane changing to overtake (c) Lane changing on a curve (d) Entering the parking area
(e) Waiting at a red light (f) Exiting the parking area (g) Turning at an intersection (h) Turning while following a vehicle} 
    \label{fig:result-qualitative}
    \vspace{-1em}
\end{figure*}

\subsection{Qualitative Result} We present 8 representative scenarios shown in Fig.\ref{fig:result-qualitative} where our DRAMA model demonstrates safe and accurate end-to-end planning results. In sub-figures (a) and (e), our planner accurately issues commands to remain static to give way to pedestrians crossing the street, regardless of the presence of explicit traffic light control. In sub-figure (a), pedestrians are crossing the street at a bend without a traffic light, while in sub-figure (e), pedestrians are crossing with traffic lights and a crosswalk. These scenarios demonstrate that our planner is able to recognize traffic lights and potential hazards, making safe planning decisions. In sub-figures (b) and (c), our planner issues lane-changing commands in response to the slow speed of the vehicle ahead. This demonstrates that our planner is capable of generating fast and complex planning operations to improve driving efficiency. Sub-figures (d) and (f) demonstrate our planner's proficiency in low-speed scenarios, specifically in maneuvering into and out of parking spaces. These examples highlight the planner's precise control and decision-making capabilities, ensuring smooth and efficient parking operations. Finally, sub-figures (g) and (h) showcase our model's planning ability in executing both right and left turns. These examples highlight the planner's adaptability in handling various traffic scenarios with precision and safety, demonstrating its comprehensive understanding of complex driving maneuvers. 

\section{Discussion and Future Work}
Due to the temporary closure of the NAVSIM Leaderboard and the limited availability of comparative solutions, we employed the public test dataset to assess both the baseline - Transfuser and our proposed methods. The baseline achieved a PDM score of 0.8483 on the NAVSIM Leaderboard; however, when tested on the public dataset, it decreased to 0.8347. Our best-performing method achieved a PDM score of 0.8548, notably surpassing the baseline on the public test dataset. The proposed multi-scale convolution contributes to DRAMA's performance sacrificing the training efficiency despite not affecting validation speed.

Given the decrease in training speed with the proposed multi-scale convolution, we will explore other powerful and efficient vision encoders. Additionally, we will also consider testing our proposed planner in real-world scenarios.

\section{Conclusion}

This work proposes a Mamba-based end-to-end motion planner named DRAMA, the first study of Mamba in motion planning for autonomous driving. Our proposed Mamba fusion and Mamba-Transformer decoder efficiently enhances the overall planning performance, and the Mamba-Transformer provides a viable alternative to traditional Transformer decoder, particularly for processing long sequences. Additionally, the feature state dropout we introduced improves the planner's robustness and can be integrated into other attention-based models, enhancing performance without increasing training or inference time. We evaluate DRAMA using the public planning dataset NAVSIM, and the results demonstrate that our method notably outperforms the baseline Transfuer with fewer parameters and lower computational costs.

%
% ---- Bibliography ----

\bibliographystyle{splncs03}
\bibliography{Drama/root.bib}
\end{document}